%% file: main.tex
\documentclass[10pt,twocolumn,letterpaper]{article}

\usepackage{cvpr}              % To produce the CAMERA-READY version
\input{preamble}

\definecolor{cvprblue}{rgb}{0.21,0.49,0.74}
\usepackage[pagebackref,breaklinks,colorlinks,citecolor=cvprblue]{hyperref}
\usepackage{bbding}
\def\paperID{*****} % *** Enter the Paper ID here
\def\confName{CVPR}
\def\confYear{2024}

\title{Hallucination Mitigation Prompts Long-term Video Understanding}

\author{Yiwei Sun\thanks{Equal contribution.} \quad Zhihang Liu$^{*}$ \quad Chuanbin Liu\thanks{Corresponding author.} \quad Bowei Pu \quad Zhihan Zhang \quad Hongtao Xie\\
University of Science and Technology of China\\
\tt\small \{syw95,liuzhihang,zhangzhihan\}@mail.ustc.edu.cn, \\
\tt\small pubowei6@gmail.com, \{liucb92,htxie\}@ustc.edu.cn
}
\begin{document}
\maketitle
\input{sec/0_abs}    
\input{sec/1_intro}
\input{sec/2_method}
\input{sec/3_exp}

{
    \small
    \bibliographystyle{ieeenat_fullname}
    \bibliography{main}
}

% WARNING: do not forget to delete the supplementary pages from your submission 
% \input{sec/X_suppl}

\end{document}

%% file: preamble.tex
\usepackage[dvipsnames]{xcolor}

%% file: sec/0_abs.tex
\begin{abstract}
    Recently, multimodal large language models have made significant advancements in video understanding tasks. However, their ability to understand unprocessed long videos is very limited, primarily due to the difficulty in supporting the enormous memory overhead. Although existing methods achieve a balance between memory and information by aggregating frames, they inevitably introduce the severe hallucination issue. To address this issue, this paper constructs a comprehensive hallucination mitigation pipeline based on existing MLLMs. Specifically, we use the CLIP Score to guide the frame sampling process with questions, selecting key frames relevant to the question. Then, We inject question information into the queries of the image Q-former to obtain more important visual features. Finally, during the answer generation stage, we utilize chain-of-thought and in-context learning techniques to explicitly control the generation of answers. It is worth mentioning that for the breakpoint mode, we found that image understanding models achieved better results than video understanding models. Therefore, we aggregated the answers from both types of models using a comparison mechanism.  Ultimately, We achieved 84.2\% and 62.9\% for the global and breakpoint modes respectively on the MovieChat dataset, surpassing the official baseline model by 29.1\% and 24.1\%. Moreover the proposed method won the third place in the CVPR LOVEU 2024 Long-Term Video Question Answering Challenge. The code is avaiable at https://github.com/lntzm/CVPR24Track-LongVideo
\end{abstract}

%% file: sec/1_intro.tex
\section{Introduction}
Long videos hold a significant position on video-sharing platforms, so building an automated long video processing system has considerable potential. Due to the enormous memory overhead, current multimodal large language models struggle to achieve perfect processing.
% \footnote{$^*$Equal contribution.}
% current research and problem
The core idea of current research is to retain more important information in the features within the memory overhead supported by existing models. MovieChat~\cite{song2024moviechat} and ChatUnivi~\cite{jin2024chat} have achieved the aggregation of information from numerous frames through weighted fusion on similar frames and dpc-knn techniques respectively. This purely aggregative approach often leads to severe hallucination issues, especially for the untrained ones. Specifically, these hallucination issues can be divided into incorrect references and fabrication. The former refers to answers that include content from the video that is irrelevant to the question, while the latter refers to answers that contain content not present in the video. They severely limit the model's understanding. Therefore, we believe mitigating hallucination is a crucial step in advancing current long video understanding tasks.

\begin{figure*}[t]
\centering
\includegraphics[width=0.85\textwidth]{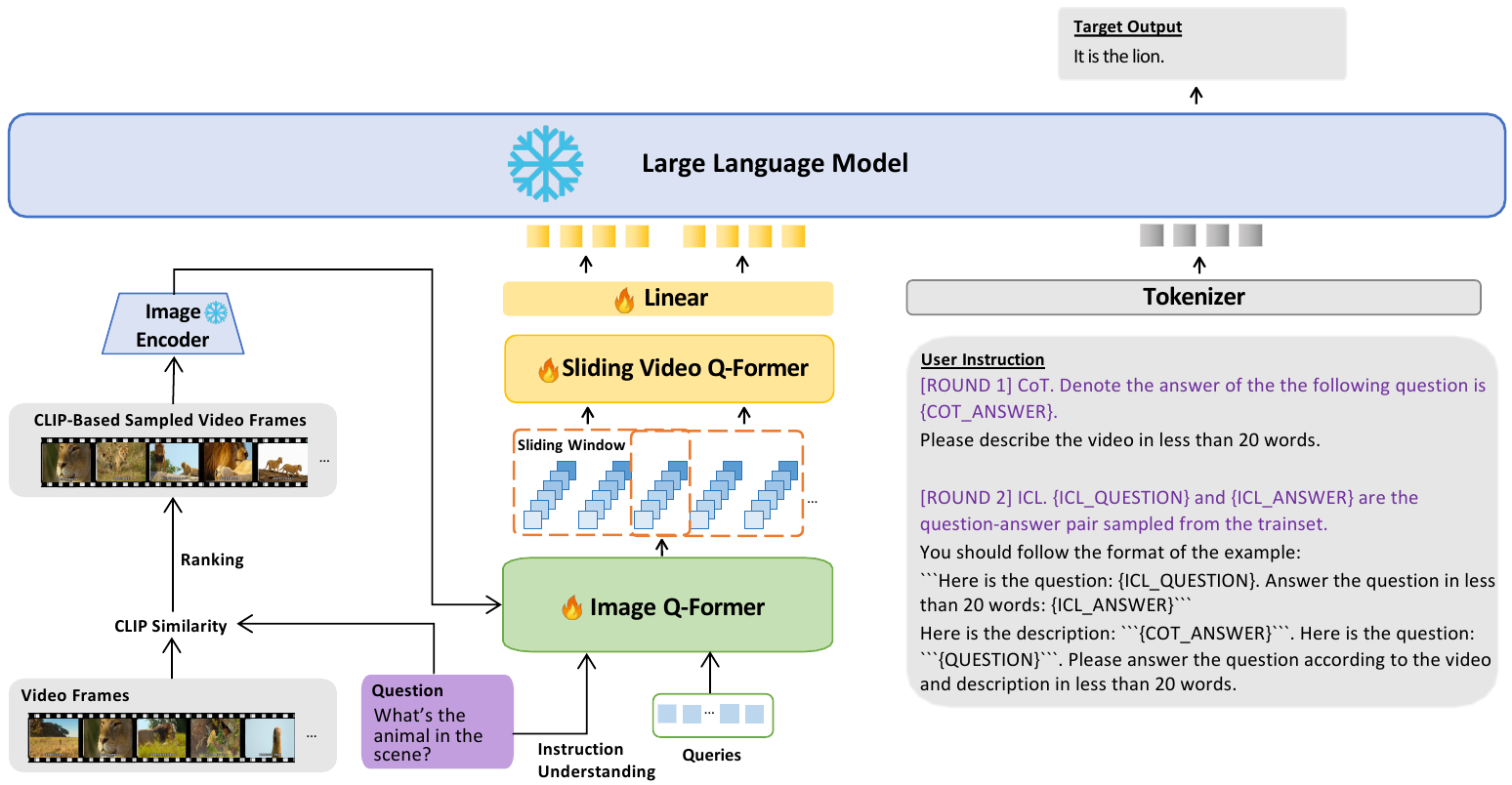}
\caption{The overall pipeline of our methodology. We introduce both training and inference technicals to get better results. For the training process, we make the visual content relevant to the instruction. For the inference process, we utilize both CoT and ICL for enhancement.}
\label{fig:method1}
\vspace{-10pt}
\end{figure*}

% our method
We propose a comprehensive hallucination mitigation pipeline on the basis of TimeChat~\cite{ren2024timechat}. To address the issue of incorrect references, we guide the information retrieval using the question. First, during the sampling stage, we calculate the CLIP~\cite{radford2021learning} Score for each frame with respect to the question and sample the top K frames with the highest scores. Then, in the feature extraction stage, following the design of InstructBLIP~\cite{dai2024instructblip}, we inject question information into the queries to guide the extraction of visual features. To address the fabrication issue, we designed a two-stage thinking process where the model first describes the image and then answers the question based on the description. Additionally, we designed an automated context example retrieval scheme to further control the answer generation.
% It is important to note that only the cross-attention module requires fine-tuning, while the rest of the components are training-free.

% vqa method
Since some question-answer pairs in  breakpoint mode do not rely on temporal information, we found that image understanding models perform better on these questions compared to video understanding models. In this paper, we used InternVL~\cite{chen2024internvl} to provide candidate answers. Then, we calculated and compared the relevance of the candidate answers to the images using CLIP. Finally, we selected the answer with the highest similarity as the final answer. 

% conclusion
The main contributions are summarized as follows:
\begin{itemize}
    \item We propose a comprehensive hallucination mitigation pipeline that effectively alleviates the hallucination issues in models.
    \item We proposed a CLIP-based comparison strategy to integrate the results of image understanding models and video understanding models which achieve better performance on the breakpoint mode.
    \item Our approach ultimately achieved 84.2\% and 62.9\% for the global and breakpoint modes respectively on the MovieChat dataset.
\end{itemize}

%% file: sec/2_method.tex
\section{Methodology}

We compare several baselines and finally choose the pre-trained TimeChat~\cite{ren2024timechat} model as our baseline. To better mitigate the hallucination issue, we introduce both training and inference technicals to get better results. As shown in Figure~\ref{fig:method1},
% for the input frames, we select the related ones based on the CLIP similarities.
for the training process, we handle the instruction-relevant visual content based on the given question in advance instead of leaving it all to the language model, 
% including CLIP-based frame sampling and instruction understanding.
% inject the instruction into the Image Q-Former, and fine-tune the sliding Video Q-Former with the time causal mask. 
For the inference process, we utilize the Chain-of-Thought and In-Context Learning to get better results to further alleviate the hallucination issue. Moreover, we introduce an image understanding model for the breakpoint mode and mix the answers together to enhance the performance.
% Since the global mode focuses on the whole video while the breakpoint mode focuses more on the current video clip, we train them separately to alleviate the gap.

\subsection{Baseline}
We choose TimeChat~\cite{ren2024timechat} as our baseline, as it uses a sliding Video Q-Former which is better for long video understanding. It uses ViT-G/14 from EVA-CLIP~\cite{sun2023eva} as the frozen image encoder, and LLaMA-2 (7B)~\cite{touvron2023llama} as the language foundation model. We fine-tune both the Image Q-Former and Video Q-Former on the MovieChat-1K train dataset while freezing the visual encoder and the LLM. Due to the gap of the data type between the global mode and the breakpoint mode, we train them separately.

\subsection{Training Process}
We notice that the existing video LLMs usually suffer from the hallucination issue when inferring in the MovieChat-1K~\cite{song2024moviechat} test dataset. We make the visual content related to the instruction in advance to mitigate the issue.

\noindent\textbf{CLIP-based Frame Sampling.} Existing methods often employ uniform sampling to extract the frames from one video. Since the video is too long, uniform sampling may lose some critical frames that are relevant to the given question. To address this problem, 
referring to the VaQuitA~\cite{wang2023vaquita},
we select the frames based on the CLIP similarity scores.
Given the input video of $L$ frames in total, we select the frames based on the similarities between the frame features and the given question. Suppose we need to sample $T$ frames for the video LLM, we first select $T/2$ frames uniformly to capture the global semantic information, and then choose another $T/2$ frames based on the CLIP scores to get the most relevant frames based on the given question. Specifically, we extract the question text using the CLIP Text Encoder, denoted as $f_q$, and extract each frame of the video using the CLIP Image Encoder as $f_\text{frame}^i$, where $i\in\{ 1,2,...,T/2 \}$, and the similarity can be calculated as:
\begin{equation}
sim\left(f_q, \ f_\text{frame}^i\right) = \frac{f_q \cdot f_\text{frame}^i}{\|f_q\|_2 \cdot \|f_\text{frame}^i\|_2}.
\end{equation}
We choose the indices of the top $T / 2$ similarities as the rest of the frames we select. Finally, we sort these frames in ascending order to get the frame features from a long video. Frames that are most related to the question will be selected by utilizing the CLIP-Score approach, improving representation learning ability.

% \subsection{Instruction Understanding and Time Region Mask for Training}
\noindent\textbf{Question-guided Frame Feature Extractor.}
% 减少幻觉，一从图像，理解问题相关的视觉信息；二从视频，局部滑窗
% In the training process, we introduce two approaches to mitigate the hallucination issue. 
To further make the visual content more related to the question, we handle the visual information with instruction understanding for the input of the Image Q-Former. Specifically, inspired by the InstructBLIP~\cite{dai2024instructblip} and TimeChat~\cite{ren2024timechat}, we add the question as the condition to fuse the visual and instruction information, concatenating the condition with the learnable queries. Since they are learned by the same self-attention, the model can extract the task-relevant visual contents, thus making the LLM understand the visual token more easily.
% \subsection{Time Region Mask for Sliding Video Q-Former}

\subsection{Inference Process}
Apart from the training process, we also utilize two straightforward strategies for the inference process, which are Chain-of-Thought (CoT) and In-Context Learning (ICL). Both of them can help to mitigate the hallucination issue and obtain better performance.

\noindent\textbf{Chain-of-Thought.}
We directly use CoT technology for our inference. Specifically, we first construct a question to let the model describe the video. Then in the second round, we use the history, which is the generated description as the input prompt, hindering the model to complete the question.

\noindent\textbf{In-Context Learning.}
ICL is another effective way to enhance the generalization ability of the model with the help of the given examples. Specifically, we calculate the similarity between the given question and the questions in the training dataset, and then select the question-answer pair with high similarity as the ICL example. With the help of the example as the prompt, the model can learn how to answer the given question better.

The final input of the prompt can refer to Figure~\ref{fig:method1}. The global mode and breakpoint mode share the same training and inference scheme.

\subsection{Comparison Strategy}
For the breakpoint mode, we think the current frame needs more attention than the current video clip and the whole video. Therefore, we introduce the InternVL~\cite{chen2024internvl} to understand the current frame and mix the results with our trained model. Specifically, we take the current frame of the breakpoint mode as the visual input, and the question as the text input, getting answers with both CoT and ICL learning. Once obtain the answer from InternVL, we mix it with our trained TimeChat by calculating the similarity between the answers and the current video clip. Suppose the answer of InternVL is $A_1$, the trained TimeChat is $A_2$ for each question, and the video clip is $V$, we use CLIP Image Encoder and Text Encoder to calculate their similarities: $sim_i = sim(\text{CLIP}_T(A_i), \text{CLIP}_I(V))$, where $i$=1,2. Then the selection strategy can be formulated as:
\begin{equation}
    A = \left\{
        \begin{aligned}
            A_1, & \ if \ sim_1 > sim_2 \\
            A_2, & \ if \ sim_1 \leq sim_2 \\
        \end{aligned}
    \right. ,
\end{equation}
and we use $A$ as the final answer of the breakpoint mode.

%% file: sec/3_exp.tex
\section{Experiments}
% 实验部分首先写baselines的比较，moviechat, timechat, chatunivi的比较，解释使用timechat的原因

% Our experiments are divided into two parts. The first part involves extensive testing of existing baseline models to confirm the baseline model used in this competition. The second part consists of ablation experiments on the proposed approach. Ultimately, we provided multiple experimental results to demonstrate the reliability of our approach.
\subsection{Baseline Comparison}

\vspace{-5pt} % 表格上方的间距
\begin{table}[!htbp]
    \centering
    \begin{tabular}{c|cccc}
        \toprule
       Baseline  & Gb Acc & Gb Sco & Bk Acc & Bk Sco\\
        \midrule
       MovieChat  & 55.3\% & 2.73 & 26.7\% & 1.41\\
       ChatUnivi  & 62.5\% & 3.07 & 32.2\% & 1.61\\
       TimeChat   & \textbf{73.8\%} & \textbf{3.58} & \textbf{46.1\%} & \textbf{2.34}\\
        \bottomrule
    \end{tabular}
    \caption{Performance of each baseline.}
    \label{tab:baseline}
\end{table}
\vspace{-5pt} % 表格上方的间距

To confirm the baseline model used, we tested the MovieChat, ChatUnivi, and TimeChat models. The results are shown in Table \ref{tab:baseline}. TimeChat surpassed MovieChat and ChatUnivi by a significant margin; therefore, we chose TimeChat as our baseline model.

\subsection{Ablation Study}

\subsubsection{Hallucination Mitigation Pipeline}
\vspace{-5pt} % 表格上方的间距
\begin{table}[!htbp]
    \centering
    \begin{tabular}{ccc|cc}
        \toprule
       CoT & CFS & ICL & Gb Acc & Bk Acc \\
        \midrule
                   &            &            &55.3\% & 26.7\% \\
        \Checkmark &            &            &62.6\% & 46.7\% \\
        \Checkmark & \Checkmark &            &69.3\% & 46.3\% \\
        \Checkmark & \Checkmark & \Checkmark &\textbf{70.5\%} & \textbf{51.9\%} \\
        \bottomrule
    \end{tabular}
    \caption{Effect of each component. Results are reported on MovieChat-test with the model MovieChat. CFS means CLIP-based Frame Sampling.}
    \label{tab:compo_1}
\end{table}
\vspace{-5pt} % 表格上方的间距

The process of designing the strategy and selecting the model was conducted simultaneously. Therefore, we performed ablation experiments on some modules using the MovieChat model. The results are shown in Table \ref{tab:compo_1}. CoT significantly improved performance in both the global and breakpoint modes. This demonstrates that such an explicit chain-of-thought process helps the model better understand the video. CLIP-based frame sampling improves the performance only in the global mode. This demonstrates that this key frame sampling strategy helps the model provide more accurate answers to questions. However, for the breakpoint mode, the impact of key frame sampling is minimal because we only provide segments near the current frame which contain few information. Finally, the ICL technique further improved performance in both the global and breakpoint modes, primarily due to the constraints on the answer format.

Table \ref{tab:compo_2} shows the effectiveness of the question-guided frame feature extractor. After incorporating this method, there was an improvement in overall performance, and the addition of other methods further enhanced the model's effectiveness.

\vspace{-5pt} % 表格上方的间距
\begin{table}[!htbp]
    \centering
    \begin{tabular}{cc|cc}
        \toprule
       QFFE & CFS+CoT+ICL & Gb Acc & Bk Acc\\
        \midrule
                   &            & 73.8\% & 46.1\% \\
        \Checkmark &            & 77.7\% & 44.7\% \\
        \Checkmark & \Checkmark & \textbf{84.2\%} & \textbf{50.6\%} \\
        \bottomrule
    \end{tabular}
    \caption{Effect of each component. Results are reported on MovieChat-test with the model TimeChat. QFFE means Question-guided Frame Feature Extractor.}
    \label{tab:compo_2}
\end{table}
\vspace{-5pt} % 表格上方的间距

\vspace{-5pt} % 表格上方的间距
\begin{table}[!htbp]
    \centering
    \begin{tabular}{c|cc}
        \toprule
       Model  & Bk Acc & Bk Sco\\
        \midrule
       TimeChat  & 50.6\%   & 2.62\\
       InternVL  & 56.5\%   & 2.80\\
       Comparison      & \textbf{62.9\%}   & \textbf{3.09}\\
        \bottomrule
    \end{tabular}
    \caption{Effect of Comparison Strategy.}
    \label{tab:rank}
\end{table}
\vspace{-5pt} % 表格上方的间距

\vspace{-5pt} % 表格上方的间距
\begin{table}[!htbp]
    \centering
    \begin{tabular}{c|cccc}
        \toprule
       Item  & Gb Acc & Gb Sco & Bk Acc & Bk Sco\\
        \midrule
       1    & 82.2\% & 3.89 & \textbf{71.1\%} & \textbf{3.55}\\
       2    & \textbf{85.7\%} & \textbf{4.22} & 62.7\% & 3.10\\
       3    & 84.2\% & 3.94 & 62.9\% & 3.09\\
       Avg  & 84.0\% & 4.01 & 65.1\% & 3.25\\
        \bottomrule
    \end{tabular}
    \caption{Mutiple results. All results are independent.}
    \label{tab:multiresults}
\end{table}
\vspace{-5pt} % 表格上方的间距

\subsubsection{Comparison Strategy}

Table \ref{tab:rank} shows the performance of the image understanding model and the video understanding model. We found that the image understanding model performs better in the breakpoint mode, indicating that some questions have a lower dependency on temporal information. To leverage the strengths of both the image understanding model and the video understanding model, we adopted a comparison strategy, which achieved a performance of 62.9\%.

\subsection{Multiple Results}

Considering the variability of evaluation methods based on large language models, we provide results from three experiments to verify the effectiveness of our approach. The results are shown in Table \ref{tab:multiresults}. By averaging the multiple results of the global mode and the breakpoint mode separately, we obtain the global accuracy 84.0\% and the breakpoint accuracy 65.1\%.

\section{Conclusion}
We propose a hallucination mitigation pipeline for addressing hallucinations in video understanding models for long video understanding tasks. The pipeline comprises three parts: CLIP-based frame sampling, question-guided frame feature extractor, and CoT\&ICL-based generation control. Our ablation experiments demonstrate that these methods are effective on MovieChat. Additionally, we introduced a CLIP-based comparison strategy to combine the advantages of video understanding models and image understanding models in the breakpoint mode. The results show that this method is effective. Overall, our approach is effective, but there is still ample room for improvement.